\documentclass{article}

\usepackage{xspace}
\usepackage{subcaption}
\usepackage{color}
\usepackage{amsmath}
\usepackage{arxiv}
\usepackage{multirow}
\usepackage[utf8]{inputenc} 
\usepackage[T1]{fontenc}    
\usepackage{hyperref}       
\usepackage{url}            
\usepackage{booktabs}       
\usepackage{amsfonts}       
\usepackage{nicefrac}       
\usepackage{microtype}      
\usepackage{lipsum}
\usepackage{fancyhdr}       
\usepackage{graphicx}       
\graphicspath{{media/}}     

\usepackage{pifont}
\usepackage{tabularx}

\newcommand{\method}{LightMotion\xspace}

\title{\method: A Light and Tuning-free Method for Simulating Camera Motion in Video Generation
}

\author{
  {Quanjian Song$^1$\footnotemark[1]~ \quad Zhihang Lin$^1$}\footnotemark[1]~ \quad Zhanpeng Zeng$^1$ \quad Ziyue Zhang$^1$ \quad Liujuan Cao$^{1}$\footnotemark[2]~ \quad Rongrong Ji$^{1}$ \\
  $^1$Key Laboratory of Multimedia Trusted Perception and Efficient Computing, \\
  Ministry of Education of China, Xiamen University, China. \\
  \texttt{\small songqj@stu.xmu.edu.cn, zhihanglin@stu.xmu.edu.cn, wiscpen@gmail.com,} \\
  \texttt{\small zhang\_zi\_yue@foxmail.com, caoliujuan@xmu.edu.cn, rrji@xmu.edu.cn} \\
  \url{https://github.com/QuanjianSong/LightMotion}
}

\begin{document}

\maketitle

\renewcommand{\thefootnote}{\fnsymbol{footnote}}
\footnotetext[1]{Equal contribution.} \footnotetext[2]{Corresponding author.}

\begin{center}
    \centering
    \vspace{-6mm}
    \includegraphics[width=1\textwidth]{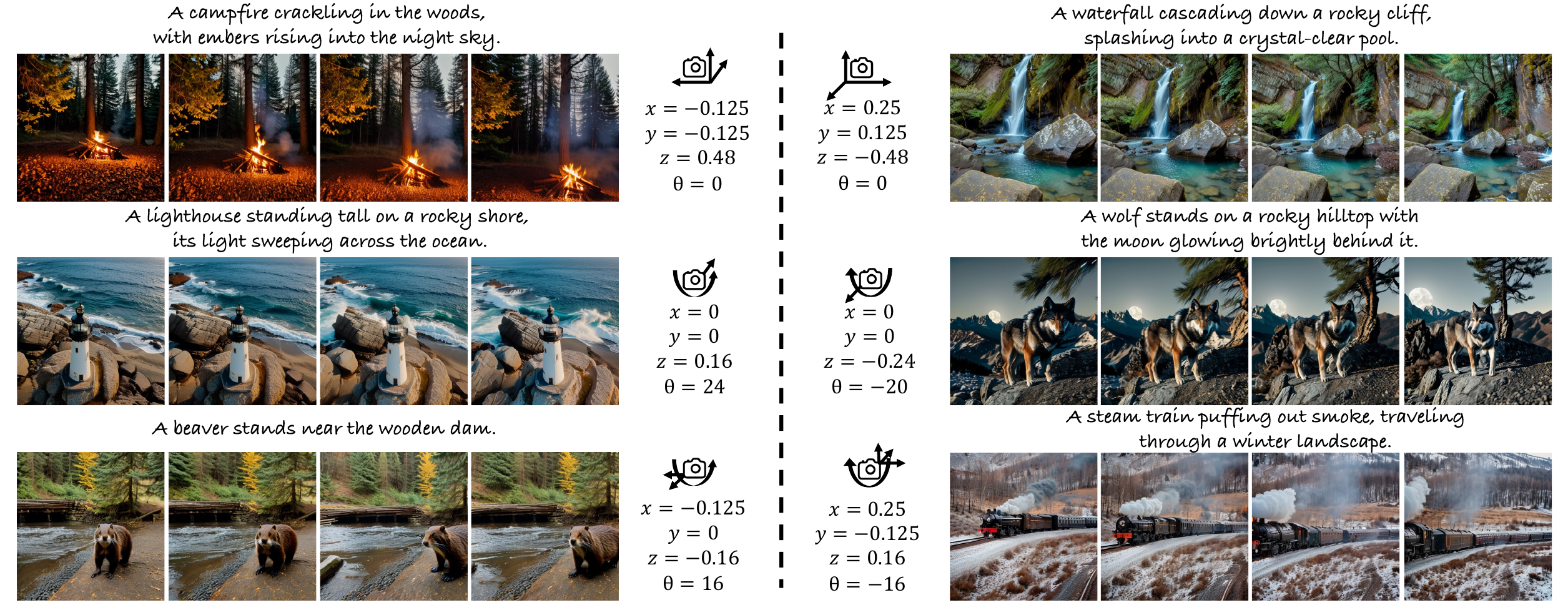}
    \vspace{-3mm}
    \captionsetup{hypcap=false}
    \captionof{figure}{ 
    LightMotion enables video generation with camera motion through user-defined parameter combinations without fine-tuning.
    }
    \label{fig:overview}
\end{center}
\begin{abstract}
Existing camera motion-controlled video generation methods face computational bottlenecks in fine-tuning and inference.
This paper proposes \method, a light and tuning-free method for simulating camera motion in video generation. 
Operating in the latent space, it eliminates additional fine-tuning, inpainting, and depth estimation, making it more streamlined than existing methods. 
The endeavors of this paper comprise:
(i) The latent space permutation operation effectively simulates various camera motions like panning, zooming, and rotation.
(ii) The latent space resampling strategy combines background-aware sampling and cross-frame alignment to accurately fill new perspectives while maintaining coherence across frames.
(iii) Our in-depth analysis shows that the permutation and resampling cause an SNR shift in latent space, leading to poor-quality generation.
To address this, we propose latent space correction, which reintroduces noise during denoising to mitigate SNR shift and enhance video generation quality.
Exhaustive experiments show that our \method outperforms existing methods, both quantitatively and qualitatively.
\end{abstract}

\keywords{Video Generation \and  Camera Motion \and Diffusion Models \and Tuning-free}
\section{Introduction}
\label{sec:intro}

%
Sora~\cite{Sora} has enabled the development of numerous open-source video generation frameworks~\cite{AnimateDiff, ModelScope, SVD, VideoCrafter}, accelerating the growth of video generation research.
Early research focuses on controllable video generation through conditions like pose maps~\cite{AnimateAnyone}, style images~\cite{UniVST}, and depth maps~\cite{VideoComposer}, achieving more precise outcomes.
However, these approaches lack specific camera controls, which remains a significant limitation.
The growing demands from films and virtual reality industries for more flexible video generation techniques render this limitation particularly problematic.

In response, researchers~\cite{MotionCtrl, CamCo, CamI2V, VideoCrafter} have recently begun exploring controllable video generation using additional camera parameters, alleviating industry needs.
Although these methods show remarkable effectiveness, they still encounter two main challenges:
(1) \textit{Expensive fine-tuning overheads.} To achieve camera motion, some methods like CameraCtrl~\cite{CameraCtrl} fine-tune an encoder that accepts camera parameters on the generic video diffusion model Animatediff~\cite{AnimateDiff} using 16 A100 GPUs.
However, the substantial computational resources constrain accessibility for most users, rendering these methods impractical for widespread adoption.
%
(2) \textit{End-to-end inference bottlenecks.} To optimize computational resources, some approaches like CamTrol~\cite{CamTrol}, use depth estimation model~\cite{Zoedepth} and inpainting model~\cite{SD} to generate new perspectives of videos in a tuning-free manner.
While effective, these extra models increase inference complexity, ultimately making end-to-end inference infeasible.
Therefore, developing a lightweight, tuning-free, and end-to-end video generation model with camera motion remains a significant challenge.

\begin{figure*}[t]
   \centering
   \includegraphics[width=0.65\textwidth] {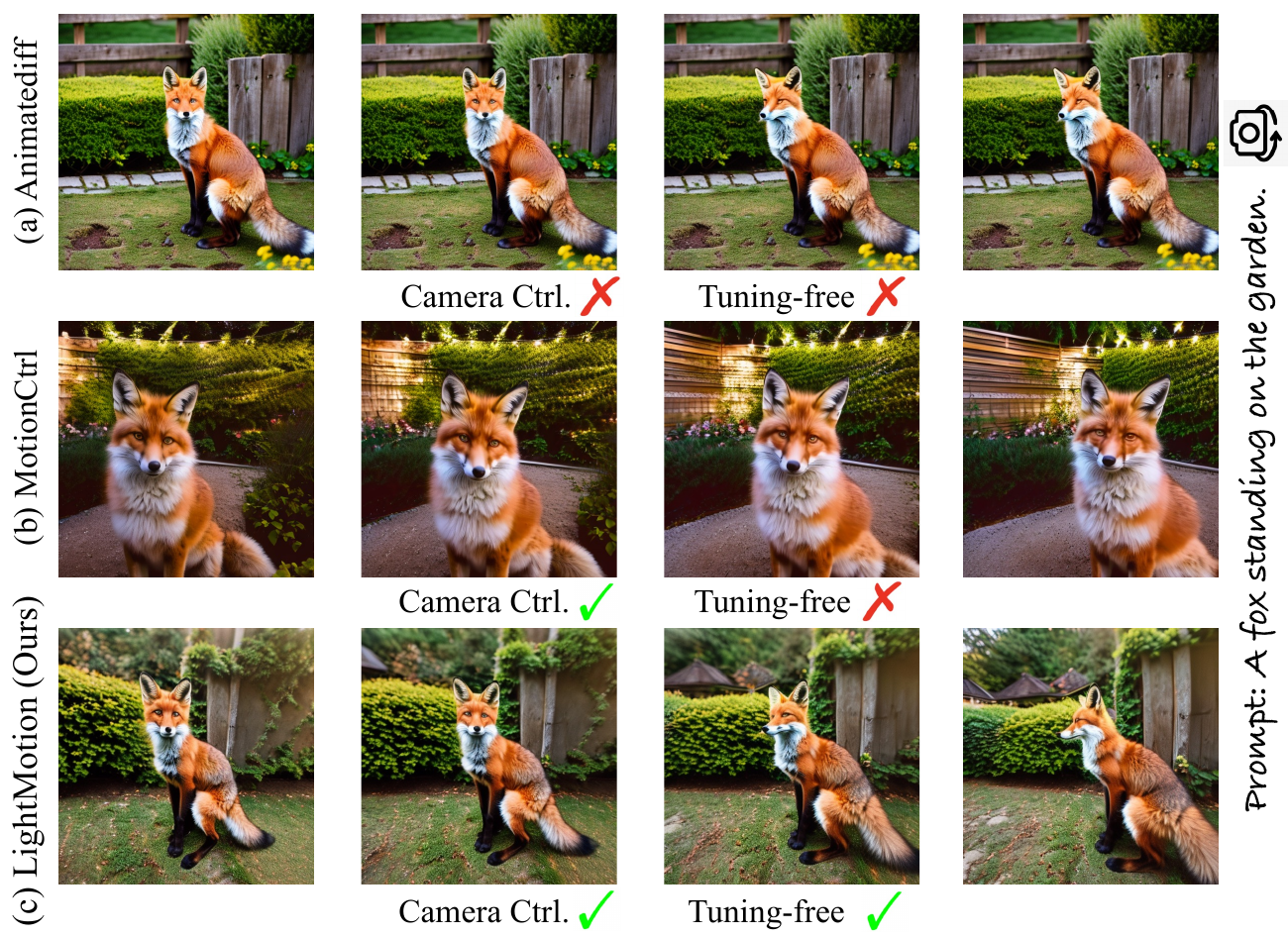}
   \caption{Comparisons with existing methods. (a)  Animatediff~\cite{AnimateDiff} produces fixed viewpoints videos. (b) MotionCtrl~\cite{MotionCtrl} fine-tunes Animatediff to achieve camera control. (c) \method allows Animatediff to simulate camera motions without fine-tuning.
   }
   \label{fig:analysis}
\end{figure*}

In this paper, we propose \textbf{\method}, a light and tuning-free method for simulating camera motion in video generation.
Operating in latent space, \method achieves end-to-end inference without additional models.
As illustrated in Figure\,\ref{fig:analysis}(a), generic video generation models like Animatediff~\cite{AnimateDiff} typically produce outputs from a fixed perspective.
To overcome this fixed perspective limitation, during denoising, we perform \textit{latent space permutation} to alter the relative order of pixels, simulating camera motions like panning, zooming, and rotation without additional fine-tuning.
%
%
Additionally, we show that the point cloud projections~\cite{VideoCrafter} are depth-independent with a fixed camera center, eliminating additional depth estimation models.

The original latent is updated through permutation, effectively simulating various camera movements.
However, some positions in the updated latent lack corresponding values due to the emergence of new perspective.
MotionBooth~\cite{MotionBooth} fills these positions by sampling pixels from the original latent based on semantic similarity. 
As shown in Figure\,\ref{fig:sampling_and_alignment}(a), this approach leads to object duplication in the new perspectives.
We empirically analyze that this issue arises from incorrect sampling of object region and introduce a \textit{background-aware sampling} strategy that utilizes the cross-attention map to achieve more accurate sampling.
Furthermore, we propose a \textit{cross-frame alignment} strategy to maintain sampling consistency across frames.

The update operation corresponding to different camera parameters has varying impacts on the generation results.
Inspired by~\cite{UpsampleGuidance}, we investigate this phenomenon from the signal-to-noise ratio (SNR).
Our in-depth analysis reveals that the tuning-free update operation induces SNR shift in the latent space, creating a gap between training and inference that degrades generation quality.
To address this, we propose a \textit{latent space correction} mechanism to uniformly correct the SNR shift.
As shown in Figure\,\ref{fig:analysis}(c), our \method achieves camera motion comparable to those of tuning-based method~\cite{MotionCtrl} depicted in Figure\,\ref{fig:analysis}(b).

Exhaustive comparisons show that \method outperforms other video generation methods with camera motion in a tuning-free manner.
%
Additionally, in Figure\,\ref{fig:overview}, \method supports various user-defined parameter combinations, further highlighting its flexibility and versatility.

\section{Related Works}
\label{sec:formatting}

\subsection{Camera Motion Video Generation via Tuning}
Video generation with camera motion has recently gained significant research interest. 
%
Early works mainly focus on fine-tuning specific datasets with camera motion.
MotionCtrl~\cite{MotionCtrl} and CameraCtrl~\cite{CameraCtrl} train an encoder to process camera parameters and inject encoded features into temporal attention layers for perspective control.
CamCo~\cite{CamCo} and CamI2V~\cite{CamI2V} leverage camera parameters to compute epipolar lines across different perspectives, constructing masks to constrain frame-to-frame attention, thereby improving the modeling of physical scene information. 
ViewCrafter~\cite{VideoCrafter} first estimates the depth map of an image and projects this image into 3D point cloud. These point clouds are projected into different perspectives, with the missing regions filled in by a fine-tuned inpainting model.


\subsection{Tuning-free Camera Motion Video Generation}
The heavy fine-tuning burden of video generation models and the scarcity of specialized datasets restrict the development of tuning-based methods.
Consequently, recent works attempt to enable base models~\cite{AnimateDiff, Lavie, Zeroscope} to perform camera-controllable video generation without additional fine-tuning.
CamTrol~\cite{CamTrol} leverages a depth estimation model~\cite{Zoedepth} to generate a 3D point cloud from an image, renders it from user-defined perspective, and uses an inpainting model~\cite{SD} to fill gaps caused by the perspectives transformation.
MotionBooth~\cite{MotionBooth} proposes a latent-shift operation to generate videos with camera motion and incorporates a random sampling strategy to fill in content for new perspectives.
However, the uncertainty in sampling can lead to object duplication in new perspectives, as shown in Figure\,\ref{fig:sampling_and_alignment}(a).
Furthermore, this method struggles to handle complex rotational and zooming motions effectively.

\section{Preliminary}

\subsection{Latent Video Diffusion Model}
The Latent Video Diffusion Model~\cite{LVD} uses an encoder $\mathcal{E}$ to map $N$ frames of video $I^{1:N}$ to the latent space $Z_1^{1:N}$. 
During training, $Z_1^{1:N}$ is noised via the diffusion process:
\begin{equation}
Z_t^{1:N}=\sqrt{\bar{\alpha}_t} \cdot Z_1^{1:N}+\sqrt{1-\bar{\alpha}_t} \cdot \varepsilon^{1:N},\quad\varepsilon^{1:N} \sim \mathcal{N}(0, I),
\label{eq:1}
\end{equation}
where $\{\alpha_t\}_{t=1}^T$ represents a predefined variance schedule, and $\bar{\alpha}_t$ is defined as the product of $\alpha_i$ from $i=1$ to $t$.

Given the noised latent $Z_t^{1:N}$, a neural network then predicts the added noise, which is supervised by the mean squared error loss and guided by the text prompt $y$.
%
Upon completion of training, the model will start with a standard Gaussian noise $Z_T^{1:N}$, progressively predicts the noise $\varepsilon_{\theta}^{1:N}$, and executes the DDIM~\cite{DDIM} schedule as follows:
\begin{equation}
\begin{array}{c}
    Z_{t-1}^{1:N}=\sqrt{\frac{\alpha_{t-1}}{\alpha_{t}}} \cdot Z_{t}^{1:N} + \left(\sqrt{\frac{1}{\alpha_{t-1}}-1}-\sqrt{\frac{1}{\alpha_{t}}-1}\right) \cdot \varepsilon_{\theta}^{1:N}.
\end{array}
\label{eq:2}
\end{equation}

After the denoising process, the decoder $\mathcal{D}$ will effectively reconstructs the final video $I'^{1:N}$ from the latent space by appling the operation $I'^{1:N}=\mathcal{D}(Z_1^{1:N})$.

\subsection{Cross-Attention Map}
In the cross-attention layers, the intermediate feature $\phi(Z_t)$ at step $t$ is mapped to $Q$ via the learnable matrix $W_Q$. Meanwhile, the text prompt $y$ is encoded by $\tau$ and mapped to $K$ and $V$ via the learnable matrices $W_K$ and $W_V$. This computes the attention between text and intermediate feature:
\begin{equation}
    \begin{aligned}
        Q &= W_Q \cdot \phi(Z_t), \quad K = W_K \cdot \tau(y), \quad V = W_V \cdot \tau(y), \\
        \mathcal{A} &= \text{Softmax}\left(\frac{Q \cdot K}{\sqrt{d}}\right), \quad \text{Attention}(Q, K, V) = \mathcal{A} \cdot V.
    \end{aligned}
\label{eq:3}
\end{equation}
where $d$ is the dimension of $Q$ and $K$, $\mathcal{A} \in \mathbb{R}^{N \times (hw) \times L}$ is the cross-attention map, $hw$ denotes the number of visual tokens, and $L$ represents the number of text tokens.
The value in the cross-attention map indicates the correlation between the text token and generated region~\cite{P2P}.

\begin{figure*}[!t]
\begin{center}
    \includegraphics[width=\textwidth]{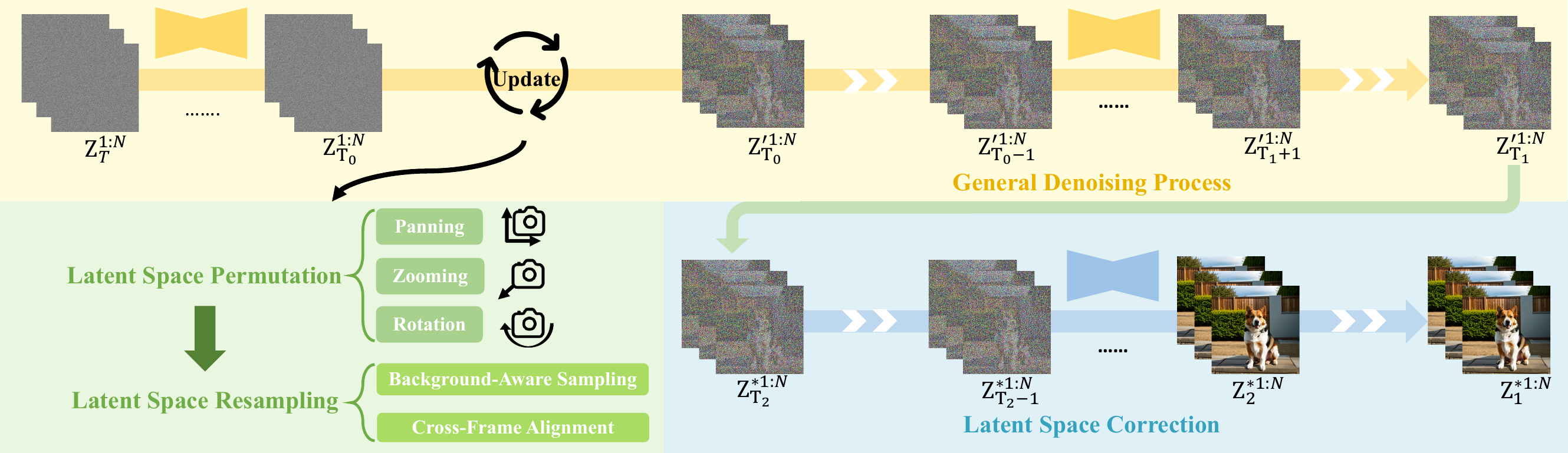}
    \caption{
The overall pipeline of \method. 
It first performs general denoising, $t = T \to T_0$.
The original latent is updated through latent space permutation and resampling.
Then, it continues denoising, $t = T_0 \to T_1$, to preserve semantic integrity and camera motion.
Next, the SNR shift is corrected by the diffusion process, $t = T_1 \to T_2$.
Finally, the general denoising continues, $t = T_2 \to 1$.
}
   \label{fig:overall pipaline}
\end{center}
\end{figure*}

\section{Method}
\subsection{Overall Pipeline}
In this section, we propose \method, a light and tuning-free method for simulating camera motion in video generation. 
The overall pipeline is illustrated in Figure\,\ref{fig:overall pipaline}.
Starting from a Gaussian noise $Z_{T}^{1:N}$, \method first denoises it until timestep $T_0$, as formulated in Eq.\,(\ref{eq:2}).
%
Then, the latent variable $Z_{T_0}^{1:N}$ is updated to $Z_{T_0}^{\prime 1:N}$ through \textit{latent space permutation} and \textit{latent space resampling}.
%
%
Next, \method continues denoising until timestep $T_1$ to preserve the semantics integrity and camera motion.
Subsequently, it performs \textit{latent space correction} to reintroduce noise from $Z_{T_1}^{\prime 1:N}$ to $Z_{T_2}^{* 1:N}$ using Eq.\,(\ref{eq:1}),  mitigating the SNR shift caused by the update operation.
Finally, \method completes the remaining denoising from timestep $T_2$ to $1$.
%

\subsection{Latent Space Permutation}
\begin{figure*}[!t]
\begin{center}
    \includegraphics[width=\textwidth]{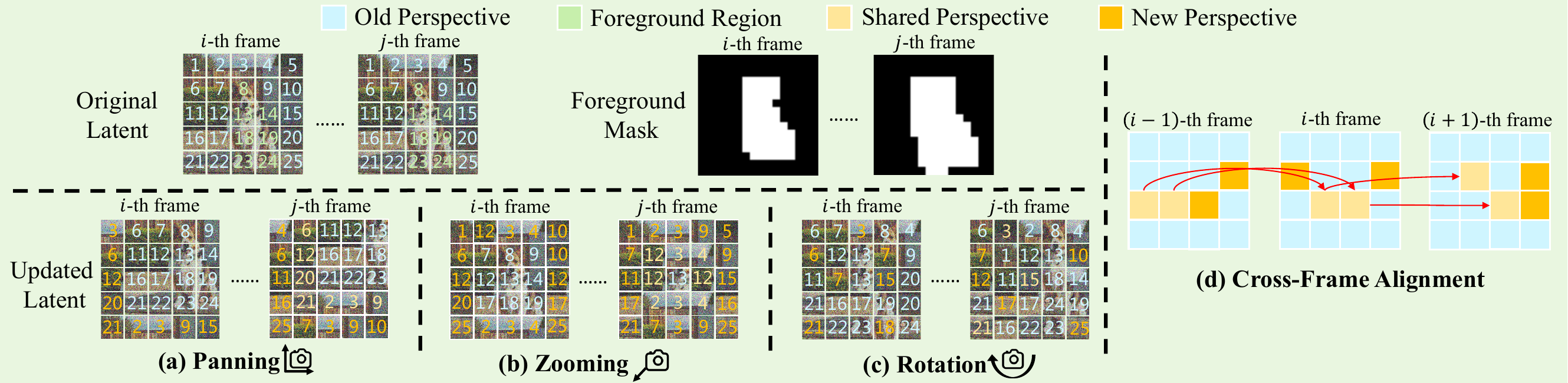}
    \caption{
An illustration of permutation and resampling in latent space with different camera motions: (a) panning, (b) zooming, and (c) rotation.
Different numbers represent distinct positions, which are rearranged into new coordinates via camera motions, followed by sampling from the old perspective to obtain the new one.
Additionally, (d) presents a toy example for the cross-frame alignment strategy.
}
   \label{fig:rearranging_sampling}
\end{center}
\end{figure*}

In reality, camera movements are inherently linked to changes in pixel positions between different frames. 
For example, in a leftward camera motion, the given pixels across different frames will roughly appear to move right.
Inspired by this, we simulate different camera motions by adjusting the relative order of pixels across frames in the latent space.
Specifically, for the pixel at coordinates $[u, v, 1]$  in the original latent $Z_{T_0}^{1:N}$, where the $1$ denotes the homogeneous coordinate, we calculate its new coordinates $[u^{\prime}, v^{\prime}, 1]$ to realistically simulate different camera motions:
\begin{equation}
[u^{\prime}, v^{\prime}, 1]^{T} = \mathcal{F}^{1:N} ([u, v, 1]^{T}),
\end{equation}
where the function $\mathcal{F}^{1:N}$ are coordinate mapping functions that vary on different camera motions, such as \textit{panning}, \textit{zooming}, and \textit{rotation}, as illustrated in Figure\,\ref{fig:rearranging_sampling}(a-c).
For brevity, coordinate rounding will be ignored below.


\noindent \textbf{Panning.}
Camera panning shifts the relative order of pixels in the latent space along the panning direction, and the coordinate mapping for the $i$-th frame is defined as:
\begin{equation}
\mathcal{F}^{i} ([u^{\prime}, v^{\prime}, 1]^{T}) := \begin{bmatrix} 1 & 0 & \frac{x \cdot w \cdot i}{N} \\ 0 & 1 & \frac{y \cdot h \cdot i}{N} \\ 0 & 0 & 1 \end{bmatrix} \cdot
\begin{bmatrix} u \\ v \\ 1 \end{bmatrix},
\end{equation}
where $x,y \in [-1, 1]$ are user-specified panning parameters following~\cite{Direct-A-Video}, which respectively define the allowable movement ranges along the $X$ and $Y$ axes as relative proportions of the width $w$ and height $h$ in the latent space.

\noindent \textbf{Zooming.}
Camera zooming is accompanied by the scaling of perspective.
Building on this, we interpolate the latent variable for each frame to simulate camera zooming, with the coordinate mapping for the $i$-th frame given by:
\begin{equation}
    \mathcal{F}^{i}([u^{\prime}, v^{\prime}, 1]^{T}) := \mathcal{T}([u, v, 1]^{T}, s^{i}),
\end{equation}
where $\mathcal{T}$ is the interpolation transformation, and the scaling factor $s^{i}$ for the $i$-th frame is given by $1 + \frac{z \cdot i}{N}$, with $z \in [-1, 1]$ as the user-specified zooming parameter like~\cite{Direct-A-Video}.

\noindent \textbf{Rotation.}
Camera rotation in latent space aligns with that in pixel space.
Based on this, we introduce the point cloud projection theory~\cite{CamTrol} for rotation modeling, with the coordinate mapping for the $i$-th frame formulated as follows:
\begin{equation}
\mathcal{F}^{i}([u^{\prime}, v^{\prime}, 1]) := K \cdot R^{i} \cdot K^{-1} \cdot \begin{bmatrix} u \\ v \\ 1 \end{bmatrix} \cdot d(u, v, 1),
\label{eq:7}
\end{equation}
where $K$ is the camera's intrinsic matrix, $R^i$ is the rotation matrix for the $i$-th frame that associated by user-specified rotation parameter $\theta$, and $d(u, v, 1)$ is associated depth information. 
Detailed camera parameter definitions and settings are provided in Appendix\,\ref{sec:definitions} and \,\ref{sec:settings}.
%
%
Under a fixed camera center assumption, the rotation coordinate projection is independent of $d(u, v, 1)$, which inherently bypasses the depth estimation.
Detailed proofs are provided in Appendix\,\ref{sec:proofs}.

Then, the pixels of updated latent $Z_{T_0}^{\prime 1:N}$ will be filled:
\begin{equation}
Z_{T_0}^{\prime 1:N}[u^{\prime}, v^{\prime}, 1] = Z_{T_0}^{1: N}[u, v, 1],
\end{equation}
where the new coordinates $[u^{\prime}, v^{\prime}, 1]$ are discarded if they exceed the latent space dimensions $h$ or $w$, as these coordinates fall outside the corresponding camera perspective.

\subsection{Latent Space Resampling}
\begin{figure*}[!t]
\begin{center}
   \includegraphics[width=0.7\columnwidth] {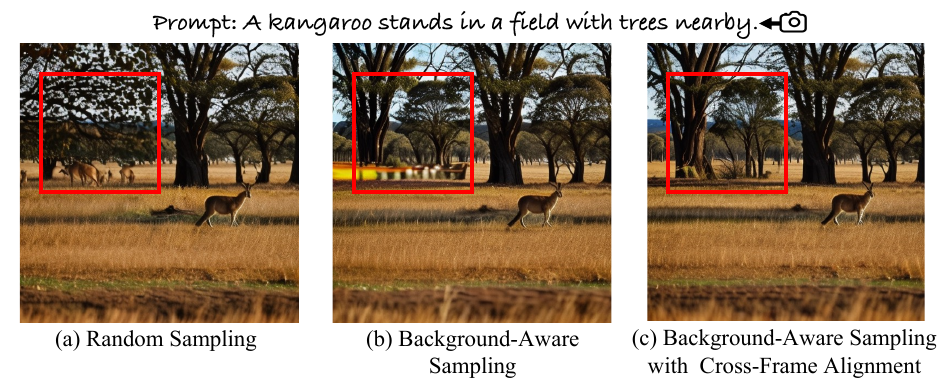}
   \caption{Results of different sampling methods: (a) Random sampling results in object repetition. (b) Background-aware
sampling results in artifacts. (c) Background-aware sampling with cross-frame alignment generates accurate results without artifacts.
}
   \label{fig:sampling_and_alignment}
\end{center}
\end{figure*}

In the previous section, we use permutations associated with different camera motions to update the latent variables.
However, some positions in the updated latent lack corresponding values.
MotionBooth~\cite{MotionBooth} fills these positions (new perspective) by randomly sampling pixels from the original latent based on semantic similarity.
Whereas, as shown in Figure\,\ref{fig:sampling_and_alignment}(a), it often produces repetitive objects.
We experimentally find that repetitive generation stems from random sampling, which incorrectly samples new pixels from the foreground region of the original latent.
%
Therefore, we propose the \textit{background-aware sampling} for more accurate sampling
, coupled with the \textit{cross-frame alignment} to ensure coherence across frames, as shown in Figure\,\ref{fig:rearranging_sampling}.


\begin{figure}[!t]
\begin{center}
   \includegraphics[width=0.7\columnwidth]{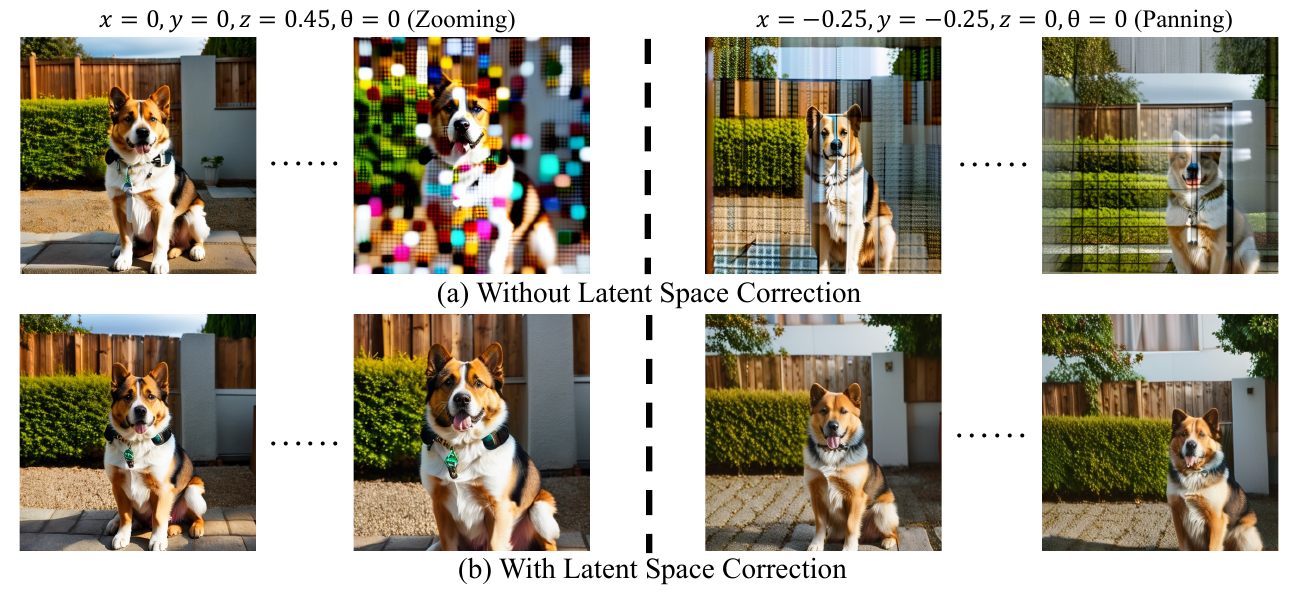}
   \caption{
(a) The update operation with different camera parameters results in poor-quality outputs.
(b) Latent space correction significantly enhances the video generation quality.
}
   \label{fig:poor-quality}
\end{center}
\end{figure}

\noindent \textbf{Background-Aware Sampling.}
Since the repetitive generation arises from incorrect sampling of foreground region in original latent, we use the cross-attention map $\mathcal{A} \in \mathbb{R}^{N \times (hw) \times L}$ formulated in Eq.\,(\ref{eq:3}) to constrain sampling region.
Specifically, we extract $\mathcal{A}$ from the last upsampling block in U-Net, during the preceding denoising process (from $T$ to $T_0$). 
Inspired by AccDiffusion~\cite{AccDiffusion}, $\mathcal{A}$ is further reshaped, binarized, and refined.
Then, we select the object token of $\mathcal{A}$ as mask $\mathcal{M} \in \mathbb{R}^{N \times h \times w}$, where $0$ denotes background region and $1$ indicates foreground region.
Finally, new pixels at $[j, k]$ in updated latent are sampled from rows or columns of the original latent's background region:
\begin{equation}
    \begin{split}
        \underset{(j,k) \in \Omega^{1:N}}{Z_{T_0}^{'1:N}[j, k]}  &= Z_t^{1:N}[j, l] \quad \textit { s.t. } \operatorname{\mathcal{M}}^{1: N}[j, l] = 0, \\
        \underset{(j,k) \in \Omega^{1:N}}{Z_{T_0}^{'1:N}[j, k]}  &= Z_t^{1:N}[l, k] \quad \textit { s.t. } \operatorname{\mathcal{M}}^{1: N}[l, k] = 0,
    \end{split}
\end{equation}
where $\Omega^{1:N}$ represents the new perspective areas across frames in the updated latent that need resampling.


\noindent \textbf{Cross-Frame Alignment.}
Background-aware sampling effectively prevents repetitive generation.
However, as shown in Figure\,\ref{fig:sampling_and_alignment}(b), the resampling region within the red box suffers from new artifacts.
We experimentally find that artifacts arise from sampling inconsistencies, where some shared pixels (shared perspective) between frames are sampled independently.
To ensure consistent sampling, we propose a cross-frame alignment strategy. 
As shown in Figure\,\ref{fig:rearranging_sampling}(d), when sampling pixels (new perspective) for the $(i+1)$-th frame, we reuse the sampled results from the $i$-th frame for shared perspective and apply independent sampling to the remaining areas.
Figure\,\ref{fig:sampling_and_alignment}(c) shows that the cross-frame alignment ensures seamless sampling propagation across frames, reducing flickering and artifacts.

\subsection{Latent Space Correction}
After the permutation and resampling in the latent space, the original latent $Z_{T_0}^{1:N}$ is updated to $Z_{T_0}^{\prime 1:N}$.
As shown in Figure\,\ref{fig:poor-quality}(a), continuing the denoising process with the updated latent as usual results in a poor-quality generation.
Inspired by~\cite{UpsampleGuidance}, we analyze this issue through the signal-to-noise ratio (SNR).
During training, the theoretical SNR of the noised latent is given by $\frac{\bar{\alpha}_t}{1 - \bar{\alpha}_t}$, as derived from Eq.\,(\ref{eq:1}).
Therefore, the model expects the input latent to have a fixed SNR at each timestep during inference.
We hypothesize that the poor-quality generation is likely caused by the permutation and resampling of the input latent, which may alter its SNR and create a gap between training and inference.
To validate this, we conduct the following two experiments.
%

\begin{figure}[!t]
\begin{center}
   \includegraphics[width=0.8\columnwidth]{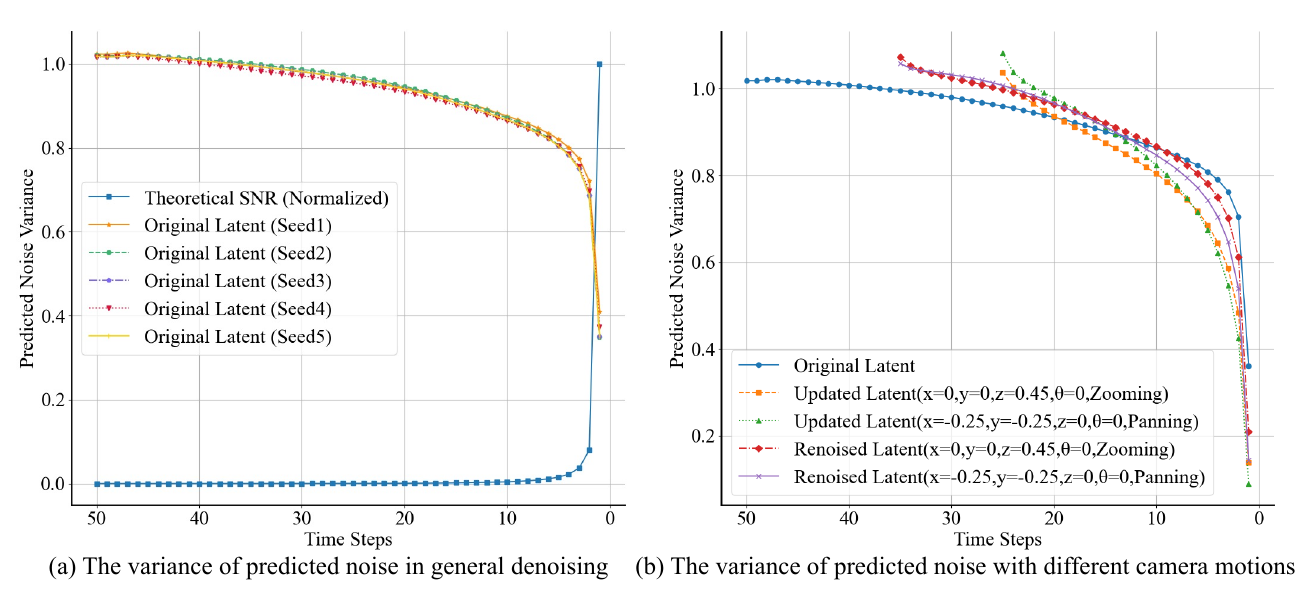}
   \caption{
(a) The variance of predicted noise is negatively correlated with the SNR of input latent.
(b) The update operation alters the variance of predicted noise, while our correction mechanism relieves this deviation.
We average the results over 1,000 samples.
}
\label{fig:SNR}
\end{center}
\end{figure}

\noindent \textbf{Exp1: Input Latent's SNR \textit{vs.} Output Noise's Variance.}
Since the update operation involves random sampling, directly estimating the SNR of the updated latent is not feasible. 
Thus, we first explore the relationship between the SNR of input latent and the variance of diffusion model's predicted noise.
As shown in Figure\,\ref{fig:SNR}(a), the variance of output noise decreases steadily, while the theoretical SNR of input latent increases gradually, during the denoising process.
This indicates that the SNR of input latent is negatively correlated with the variance of output noise.
In addition, the variances of predicted noise across different random seeds show high consistency, implying that the model outputs noise with a specific variance for the input with a given SNR.
In this way, the SNR change of input latent can be inferred from the variance change of predicted noise.

\noindent \textbf{Exp2: Update Operation Alters Input Latent's SNR.}
Based on the previous conclusion, we analyze the impact of the update operation on the input latent's SNR by examining the variance of the predicted noise, as shown in Figure\,\ref{fig:SNR}(b).
At timestep $T_0=25$, the updated latent outputs significantly higher noise variance, confirming that the update operation alters its SNR.
This change creates a gap between training and inference, hindering accurate noise prediction in subsequent denoising.
Additionally, the effect of update operation on the input latent's SNR depends on camera parameters. 
Since camera movements cause old perspectives to disappear and new ones to emerge, larger SNR changes occur when more new perspectives are introduced.

%
%
Extensive experiments show that reintroducing noise into the latent during denoising helps mitigate the SNR shift caused by the update operation, resulting in unified correction.
In this way, we continue denoising until reaching $Z_{T_1}^{'1:N}$ and then reintroduce noise as formulated in Eq.\,(\ref{eq:1}) to obtain ${Z^{*}}_{T_2}^{1:N}$.
Finally, we gradually resume the denoising process from $T_2$ to $1$.
As shown in Figure\,\ref{fig:SNR}, the output noise variance of renoised latent is closer to that of the original latent than the updated latent, indicating that this correction strategy effectively mitigates the SNR shift.
In Figure\,\ref{fig:poor-quality}(b), our latent space correction greatly enhances video generation quality across different camera parameters.

\section{Experiments}
\subsection{Experimental Settings}
\noindent \textbf{Implementation Details.}
\method builds on a widely-used T2V framework Animatediff-V2~\cite{AnimateDiff}, which is trained on 16-frame sequences at a resolution of $512 \times 512$.
Integration with other frameworks is provided in Appendix\,\ref{sec:compatibility}.
The hyper-parameters used in \method are set as follows: $T_0 = 0.5T$, $T_1 = 1$, $T_2 = 0.7T$, and $T = 50$.
\method can generate videos with camera motion in minutes on one NVIDIA GeForce RTX $3090$ GPU, without fine-tuning.

\noindent \textbf{Datasets Details.}
Following recent work~\cite{MotionBooth,CamTrol}, we evaluate the camera motion of different methods by collecting $100$ diverse prompts, each paired with $16$ distinct camera motions, ultimately resulting in $1,600$ prompt-motion pairs.
This dataset evaluates the videos generated for different objects in various scenarios and camera motions.

\noindent \textbf{Baselines.}
We compare our \method with five camera motion control baselines: Animatediff with LoRA~\cite{AnimateDiff}, Direct-A-Video~\cite{Direct-A-Video}, CameraCtrl~\cite{CameraCtrl}, MotionCtrl~\cite{MotionCtrl} and MotionBooth~\cite{MotionBooth}. Note that Motionbooth and our \method are tuning-free, while others need fine-tuning.

\begin{table*}[!t]
\caption{Quantitative experiments with the full dataset for existing methods.
The best result is in \textbf{bold} while the second-best is in \underline{underline}.
}
    \centering
    \resizebox{\textwidth}{!}{ 
    \begin{tabular}{lccccccccc}
        \toprule
        \multirow{2}{*}{\centering Methods} & \multicolumn{3}{c}{Generation Quality} & \multicolumn{3}{c}{Camera Controllability} & \multicolumn{2}{c}{User Study} & \multirow{2}{*}{\centering Tuning-free} \\ 
        \cmidrule(lr){2-4} \cmidrule(lr){5-7} \cmidrule(lr){8-9}
         & FVD$\downarrow$ & CLIP-F $\uparrow$ & CLIP-T $\uparrow$ & Pan.-Error$\downarrow$ & Zoom.-Error$\downarrow$ & Rot.-Error$\downarrow$ & Quality$\uparrow$ & Controllability$\uparrow$ & \\ \midrule
        Animatediff~\cite{AnimateDiff}         & 5645.6 & 98.28 & 32.16 & \underline{0.0873} & 0.1913 &    -   & \underline{79.93} & 86.37 & \ding{55} \\ 
        Direct-A-Video~\cite{Direct-A-Video}           & \underline{5340.2} & 97.87 & 31.16 & 0.3447 & 0.1742 &    -   & 60.53 & 59.89 & \ding{55} \\ 
        CameraCtrl~\cite{CameraCtrl}                     & 5495.6 & 98.56 & \underline{33.48} & 0.1896 & 0.2534 & 0.4211 & 48.56 & 55.29 & \ding{55} \\ 
        MotionCtrl~\cite{MotionCtrl}                     & 5498.2 & \underline{98.59} & 31.19 & 0.2298 & \underline{0.1599} & \underline{0.2871} & 79.88 & \underline{88.39} & \ding{55} \\ 
        MotionBooth~\cite{MotionBooth}                   & 5505.3 & 97.86 & 33.18 & 0.1386 &    -   &    -   & 64.50 & 66.47 & \ding{51} \\ 
        \method(Ours)                                   & \textbf{5329.5} & \textbf{98.62} & \textbf{33.56} & \textbf{0.0532} & \textbf{0.1590} & \textbf{0.2351} & \textbf{88.55} & \textbf{88.70} & \ding{51} \\ 
        \bottomrule
    \end{tabular}
    }
    \label{tab:method_comparison}
\end{table*}

\begin{table*}[!t]
\caption{Overall performance evaluated by GPT-4o~\cite{GPT4} compared with existing methods. 
The \textbf{bold} emphasizes the best result, and the \underline{underline} highlights the second-best result.
}
    \centering
    \resizebox{0.6\textwidth}{!}{ 
    \begin{tabular}{lcccc}
        \toprule
        Methods & Quality$\uparrow$ & Coherence$\uparrow$ & Controllability$\uparrow$ & Avg$\uparrow$ \\
        \midrule
        Animatediff~\cite{AnimateDiff} & \underline{91.0} & 85.3 & 85.0 & 87.1 \\
        Direct-A-Video~\cite{Direct-A-Video} & 85.8 & 84.7 & 74.8 & 81.8 \\
        CameraCtrl~\cite{CameraCtrl} & 86.1 & 83.1 & 80.1 & 83.1 \\
        MotionCtrl~\cite{MotionCtrl} & 90.8 & \underline{85.8} & \textbf{87.2} & \underline{87.9} \\
        MotionBooth~\cite{MotionBooth} & 82.5 & 82.5 & 70.0 & 78.3 \\
        \method(Ours) & \textbf{92.1} & \textbf{85.9} & \underline{86.3} & \textbf{88.1} \\
        \bottomrule
    \end{tabular}
    }
    \label{tab:gpt_comparison}
\end{table*}

\subsection{Quantitative Comparison}
%
We evaluate the quality of generated videos with camera motion across four aspects: \textit{generation quality}, \textit{camera controllability}, \textit{user study}, and \textit{GPT-4o evaluation}.

\noindent \textbf{Generation Quality.}
To evaluate the generation quality, we use three metrics: Fréchet Video Distance (FVD)~\cite{FVD}, CLIP text-image similarity (CLIP-T)~\cite{clip}, and CLIP frame-by-frame similarity (CLIP-F)~\cite{clip}.
These metrics measure the visual quality, text-video alignment, and semantic coherence across frames. 
As shown in Table\,\ref{tab:method_comparison}, our \method outperforms all baselines in FVD, CLIP-T, and CLIP-F, showing significant advantages in the visual quality, text-video alignment, and semantic coherence across frames.

\noindent \textbf{Camera Controllability.}
Camera controllability is assessed by errors in panning, zooming, and rotation~\cite{VideoFlow,Dust3R}, referred as Pan.-Error, Zoom.-Error, and Rot.-Error.
%
All methods are evaluated based on the camera motions they support.
%
As shown in Table\,\ref{tab:method_comparison}, our \method supports all three motions and achieves the best performance on Pan.-Error, Zoom.-Error, and Rot.-Error, showing its outstanding camera controllability in a tunning-free manner.
%


\noindent \textbf{User Study.}
In addition, we invite $100$ participants to score the quality and controllability of videos generated by different methods, with a maximum score of $100$.
As shown in Table\,\ref{tab:method_comparison}, our \method outperforms other baselines in both quality and controllability based on user preferences.
Further details are provided in Appendix\,\ref{sec:user_study}.

\begin{table*}[!t]
    \caption{The ablation study for the three core components. 
    It is conducted on the half of dataset, with the best results highlighted in \textbf{bold}.
    The ``B.A.Sampling'' represents background-aware sampling while the ``C.F.Alignment'' denotes cross-frame aligment.
    }
    \centering
    \resizebox{\textwidth}{!}{ 
    \begin{tabular}{ccc|cccccc}
        \toprule
        \multicolumn{3}{c|}{Different Components} & \multicolumn{3}{c}{Generation Quality} & \multicolumn{3}{c}{Camera Controllability} \\ 
        \cmidrule(lr){1-3} \cmidrule(lr){4-6} \cmidrule(lr){7-9}
        B.A.Sampling & C.F.Alignment & Latent Space Correction & FVD $\downarrow$ & CLIP-F $\uparrow$ & CLIP-T $\uparrow$ & Pan.-Error $\downarrow$ & Zoom.-Error $\downarrow$ & Rot.-Error $\downarrow$ \\ \midrule
        \ding{55} & \ding{55} & \ding{55} & 5198.9          & 96.43 & 32.58   & 0.0305 & 0.3078 & 0.7619 \\ 
        \ding{51} & \ding{55} & \ding{55} & 4937.7          & 96.47 & 32.64   & 0.0256 & 0.2813 & 0.7821  \\ 
        \ding{55} & \ding{51} & \ding{55} & 5181.6          & 96.62 & 32.59   & 0.0346 & 0.3272 & 0.7939  \\ 
        \ding{55} & \ding{55} & \ding{51} & 6157.9          & 98.35 & 33.20   & 0.0548 & 0.1878 & 0.2461  \\ 
        \ding{51} & \ding{51} & \ding{55} & \textbf{4907.5} & 96.75 & 32.65   & 0.0232 & 0.2577 & 0.8021  \\ 
        \ding{51} & \ding{55} & \ding{51} & 5417.2 & 98.34 & 33.23  & 0.0234 & 0.1844 & 0.2432  \\
        \ding{55} & \ding{51} & \ding{51} & 5751.2 & \textbf{98.36} &  33.21  & 0.0305 & 0.1828 & 0.2379  \\
        \ding{51} & \ding{51} & \ding{51} & 5226.1 & \textbf{98.36} & \textbf{33.27} & \textbf{0.0179} & \textbf{0.1756} & \textbf{0.2357}  \\ 
        \bottomrule
    \end{tabular}
    }
    \label{tab:component_ablation}
\end{table*}

\noindent \textbf{GPT-4o Evaluation.}
Futhermore, we use GPT-4o~\cite{GPT4} to assess video quality, content coherence, and camera controllability across methods.
In Table\,\ref{tab:gpt_comparison}, our \method performs best in quality and coherence, slightly behind in controllability, yet delivers superior average performance.
Further details are provided in Appendix\,\ref{sec:gpt4o}.

\begin{figure*}[htbp]
\begin{center}
   \includegraphics[width=\textwidth]{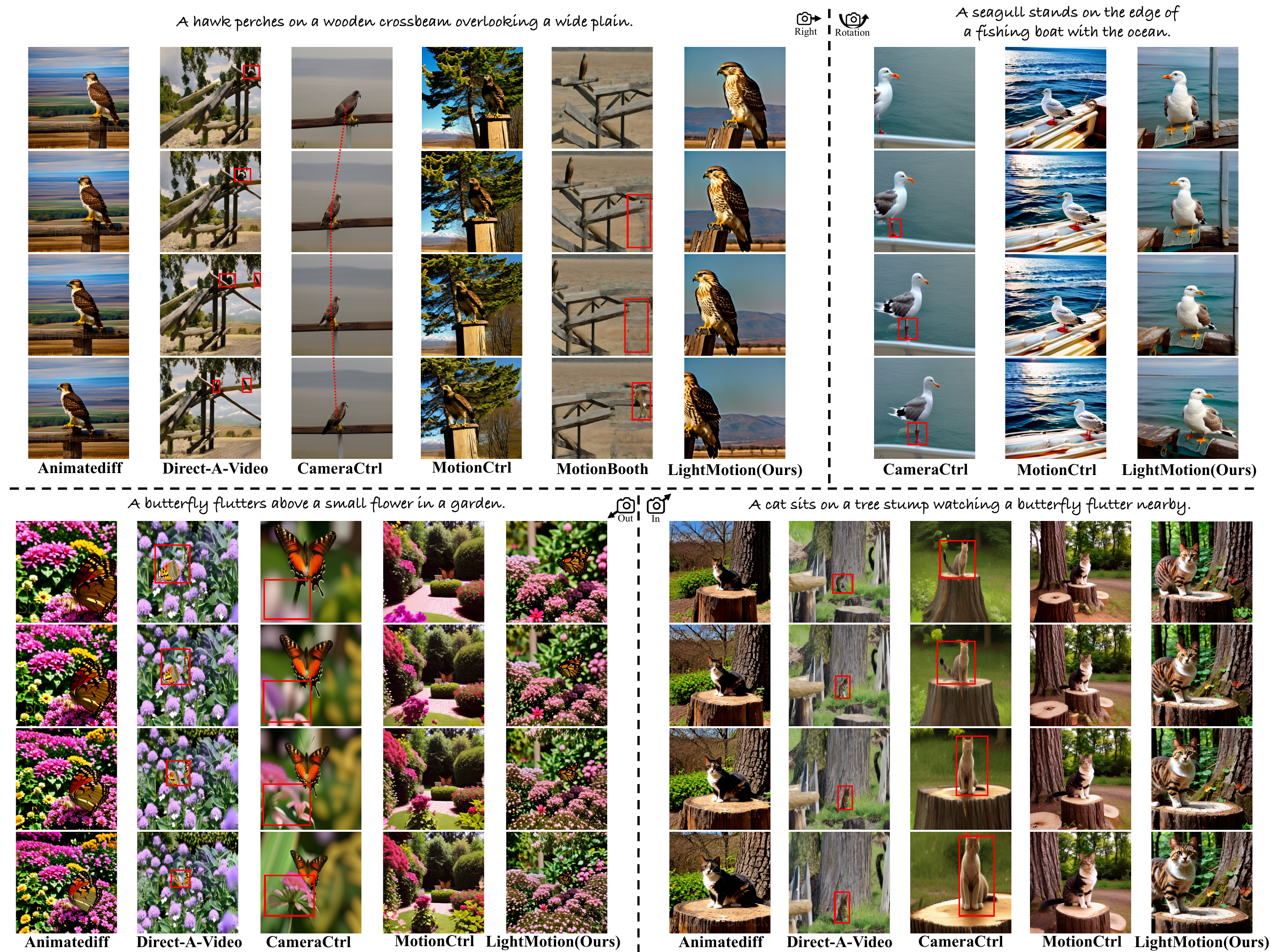}
   \caption{Qualitative comparisons with existing methods. For each camera motion, we compare only those methods that support it.
   %
   }
   \label{fig:qualitative_comparison}
\end{center}
\end{figure*}

\subsection{Qualitative Comparison}
Figure\,\ref{fig:qualitative_comparison} presents qualitative comparisons that show only the camera motions supported by each method, with the additional results are provided in Appendix\,\ref{sec:additional_qualitative} and \,\ref{sec:combinations}.
The results highlight our \method's superior performance across all camera motions, outperforming existing methods in generation quality, controllability, and coherence.
For \textit{camera panning}, MotionBooth~\cite{MotionBooth} and Direct-A-Video~\cite{Direct-A-Video} suffer from object repetition, while CameraCtrl~\cite{CameraCtrl} is insensitive to long-distance movements.
For \textit{camera zooming}, Direct-A-Video introduces artifacts that distort the object's semantics, while CameraCtrl causes inconsistencies between frames.
For camera rotation, CameraCtrl suffers from significant degradation in visual quality.

\subsection{Ablation Studies}
In this section, we validate the effectiveness of the three core components proposed in our \method, followed by an analysis of the key hyper-parameter settings.
\subsubsection{Ablation Study on Core Components}
We investigate three core components proposed in our \method: \textit{background-aware sampling}, \textit{cross-frame alignment}, and \textit{latent space correction}, validating them across eight variants.
Due to the limitations of computing resources, each variant is validated on the half of dataset, as shown in Table\,\ref{tab:component_ablation}.
In particular, the background-aware sampling and cross-frame alignment strategies collaborate to significantly enhance both generation quality and coherence across frames.
Meanwhile, the latent space correction mechanism greatly enhances camera controllability with only a minor trade-off in FVD.
%
Overall, the synergy of these three components effectively enhances stability and achieves an optimal balance in performance.

\begin{figure*}[!t]
\begin{center}
   \includegraphics[width=\textwidth]{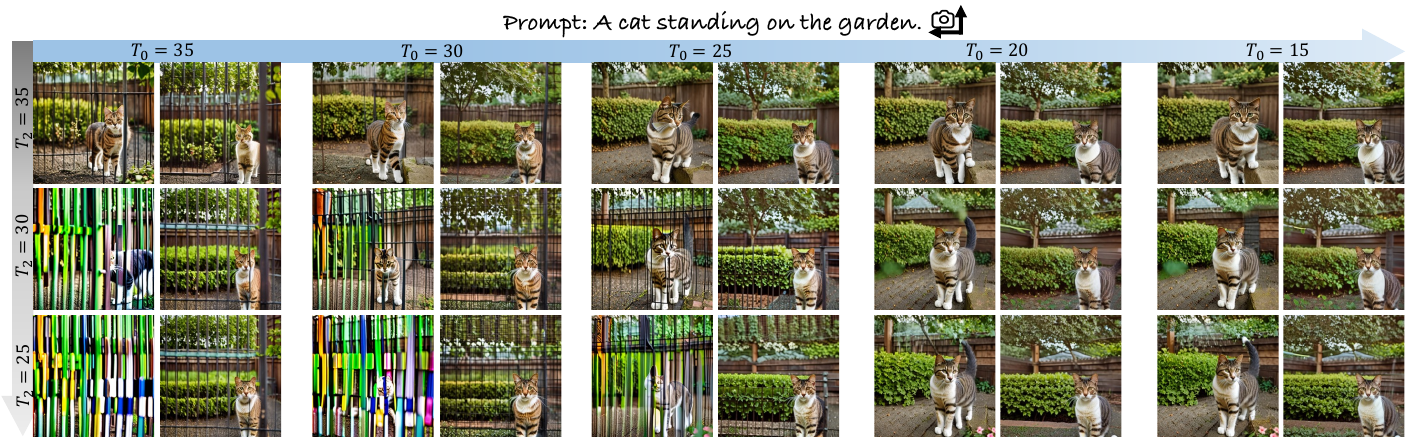}
   \caption{The ablation study of different hyper-parameters $T_0$ and $T_2$. It clearly reveals that both the updated timestep $T_0$ for latent and the noised timestep $T_2$ for correction play crucial roles in determining the results, significantly impacting overall performance.
   }
   \label{fig:hyper-parameters}
\end{center}
\end{figure*}

\begin{figure}[!t]
   \centering
   \includegraphics[width=0.6\linewidth] {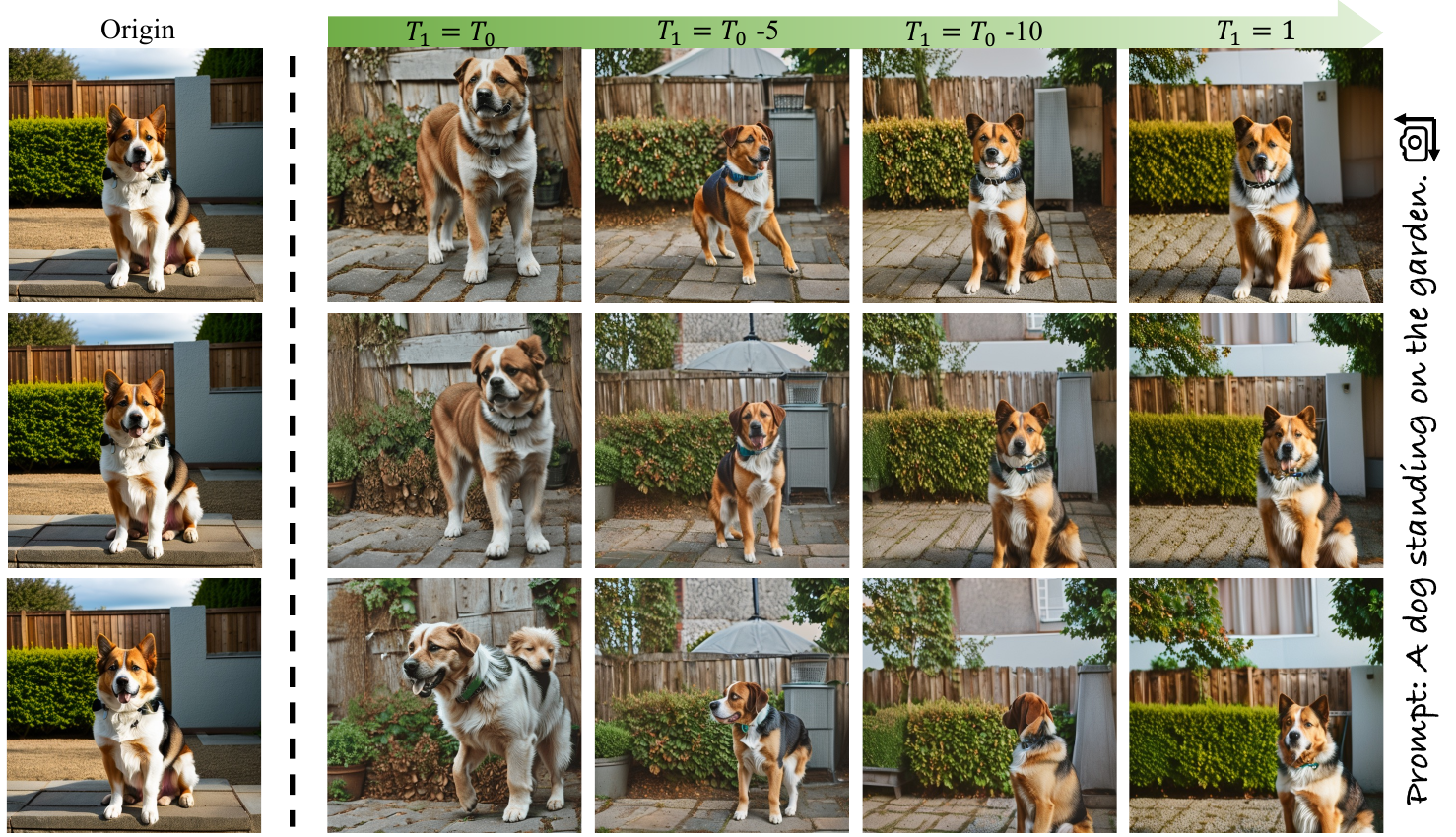}
   \caption{The ablation study of hyper-parameter $T_1$. 
   Introducing noise early disrupts semantic integrity and camera motions.
   }
   \label{fig:hyper-parameters2}
\end{figure}

\subsubsection{Ablation Study on Hyper-Parameters}
Additionally, we investigate the setting of three hyper-parameters: $T_0$, $T_1$, and $T_2$.
First, the grid search is used to explore the optimal combination, the timestep $T_0$ for update operation and the timestep $T_2$ for correction mechanism, with details provided in Figure\,\ref{fig:hyper-parameters}.
Early update operation has a low signal-to-noise ratio and causes artifacts, while later ones offer richer semantics but may lead to mismatches in new perspectives.
In terms of latent space correction, a short noised timestep fails to correct the SNR shift, while a longer noised step increases inference time.
Furthermore, we investigate the impact of introducing noise at different timestep $T_1$, with the visual results shown in Figure\,\ref{fig:hyper-parameters2}.
Introducing noise too early after the update disrupts the object's semantic integrity and yields minimal camera motions simulated by permutation.
To match the training process, we set $T_1 = 1$ to first obtain nearly clean $Z_1^{'1:N}$ and then perform the subsequent correction process.

\section{Limitations and Future Work}
Our \method also has some limitations: 
(i) The Latent space permutation and resampling strategies perform poorly in cases involving rapid camera motion with numerous new perspectives.
(ii) While effective, the latent space correction mechanism incurs additional inference time.

In this way, more effective strategies for modeling high-speed camera movements and more efficient mechanisms for correcting the SNR shift can be explored in future work.

\section{Conclusion}
We propose \method, a light and tuning-free approach that enhances video generation with camera motion.
\method operates in the latent space, eliminating the need for additional inpainting or depth estimation, achieving end-to-end inference.
Our contributions include: 
(1) Latent space permutation effectively simulates various camera motions.
(2) Latent space resampling incorporates background-aware sampling and cross-frame alignment, capturing new perspectives while maintaining frame consistency.
(3) Latent space correction mitigates the SNR shift caused by permutation and resampling, enhancing video quality.
Exhaustive experiments show that our method surpasses existing methods in both quantitative metrics and qualitative evaluations.

{
    \bibliographystyle{unsrt}  
    \bibliography{main}  
}
\clearpage
\appendix

\section{Detailed Experimental Settings}
\subsection{Camera Parameter Definitions}
\label{sec:definitions}
%
Unlike traditional camera-controlled video generation models, our \method eliminates the need for users to input technical camera parameters such as intrinsic, rotation, or translation matrices.
Instead, we streamline the input parameters without requiring knowledge of camera geometry, thus lowering the barrier for non-professional users.
Specifically, We only require users to input four camera parameters: $x,y,z$, and $\theta$. 
By combining these parameters, we can simulate various camera motions in the real world.
%

Following previous work, Direct-A-Video ~\cite{Direct-A-Video}, we define the parameters as follows:
$x$ represents the $X$-pan ratio, defined as the total horizontal shift of the frame center from the first to the last frame related to the frame width, with $x>0$ indicating the panning rightward. 
$y$ denotes the $Y$-pan ratio, which indicates the total vertical shift of the frame center related to the frame height, with $y > 0$ indicating the panning downward. 
$z$ refers to the $Z$-pan zooming ratio, defined as the scaling factor between the first and last frame, with $z>0$ indicating zooming-in.

Different from Direct-A-Video, we additionally model the camera rotation and define relative parameters.
We model rotation using point cloud projection theory, which primarily involves camera intrinsic parameters $K$, rotation matrices $R^{i}$, and depth information $d(u,v,1)$.
Here, we ignore $d(u,v,1)$ (which will be discussed in the following section) and only consider the settings of $K$ and $R^{i}$ (rotation about the Y-Axis, the same applies to other cases):
\begin{equation}
K=\left(\begin{array}{ccc}
f_{x} & 0 & u_{0} \\
0 & f_{y} & v_{0} \\
0 & 0 & 1
\end{array}\right), R^{i} = \left(\begin{array}{ccc}
\cos{\gamma^{i}} & 0 & \sin{\gamma^{i}} \\
0 & 1& 0 \\
-\sin{\gamma^{i}} & 0 & \cos{\gamma^{i}}
\end{array}\right).
\end{equation}

Similar to pixel space, the camera's optical center $(u_0, v_0)$ are positioned at the center of the latent space, $(\frac{h}{2}, \frac{w}{2})$.
However, the focal lengths $(f_x, f_y)$ in the latent space do not have physical significance. 
Through extensive experimentation, we found that $f_x = f_y = 15$ yields effective results in the latent space.
Regarding the rotation matrix $R^{i}$ for the $i$-th frame, relative angles $\gamma^{i}$ are defined as $\frac{2 \cdot \theta}{N} \cdot (i - N)$, with $\gamma^{i}$ ranging from $-\theta$ to $\theta$ across $N$ frames. Here, $\theta$ is the user-defined rotation parameter, and $\theta > 0$ indicates counterclockwise rotation.

\subsection{Camera Parameter Settings}
\label{sec:settings}
Since not all methods support every type of camera motion, we define 16 distinct camera motion scenarios, including $8$ panning, $4$ zooming, and $4$ rotation sequences, to assess the performance of each model on the respective motion types.
Following, we will provide a detailed description of the parameter settings for these camera motions.

For panning, we define $4$ motion types, including leftward, rightward, upward, and downward movements, each with two variations: small-scale and large-scale.
Small-scale panning shifts the frame from first to last, covering $25\%$ of the frame width, while large-scale panning covering $50\%$.
Parameter settings are detailed in Table\,\ref{tab:panning_paras}.

\begin{table*}[!ht]
\centering
    \caption{Camera panning parameter settings.}
    \resizebox{0.8\textwidth}{!}{ 
    \begin{tabularx}{\linewidth}{@{}l>{\centering\arraybackslash}X@{}}
    \toprule
    Camera Motion & Parameter Settings \\
    \midrule
    Leftward  (small-scale)  & $x=-0.25,y=0.00$  \\
    Leftward  (large-scale)  & $x=-0.50,y=0.00$  \\
    Rightward (small-scale)  & $x=0.25,y=0.00$  \\
    Rightward (large-scale)  & $x=0.50,y=0.00$  \\
    Upward    (small-scale)  & $x=0.00,y=-0.25$  \\
    Upward    (large-scale)  & $x=0.00,y=-0.50$  \\
    Downward  (small-scale)  & $x=0.00,y=0.25$  \\
    Downward  (large-scale)  & $x=0.00,y=0.50$  \\
    \bottomrule
    \label{tab:panning_paras}
    \end{tabularx}
    }
\end{table*}

For zooming, we define two motion types: zooming-in and zooming-out, each having small-scale and large-scale variations.
Small-scale zooming scales the frame from first to last, covering $24\%$ of the frame size, while large-scale spanning $48\%$. 
Parameter settings are detailed in Table\,\ref{tab:zooming_paras}.

\begin{table}[!ht]
\centering
    \caption{Camera zooming parameter settings.}
    \resizebox{0.8\textwidth}{!}{ 
    \begin{tabularx}{\linewidth}{@{}l>{\centering\arraybackslash}X@{}}
    \toprule
    Camera Motion & Parameter Settings \\
    \midrule
    Zooming-in  (small-scale)   &  $z=0.24$  \\
    Zooming-in  (large-scale)   &  $z=0.48$  \\
    Zooming-out (small-scale)   &  $z=-0.24$  \\
    Zooming-out  (large-scale)  &  $z=-0.48$  \\
    \bottomrule
    \label{tab:zooming_paras}
    \end{tabularx}
    }
\end{table}

For rotation, we also define two motion types: counterclockwise rotation and clockwise rotation, each having small-scale and large-scale variations.
Small-scale rotation rotates the frame from first to last, ranging from $-\theta$ to $\theta$ where $\theta = 8$, while large-scale rotation uses $\theta = 16$.
Parameter settings are detailed in Table\,\ref{tab:rotation_paras}.

\begin{table}[!ht]
\centering
    \caption{Camera rotation parameter settings. ``CCW.'' represents the counterclockwise while ``CW.'' renotes the clockwise.}
    \resizebox{0.8\textwidth}{!}{ 
    \begin{tabularx}{\linewidth}{@{}l>{\centering\arraybackslash}X@{}}
    \toprule
    Camera Motion & Parameter Settings \\
    \midrule
    CCW. rotation  (small-scale)   &  $\theta=8$  \\
    CCW. rotation  (large-scale)   &  $\theta=16$  \\
    CW. rotation (small-scale)   &  $\theta=-8$  \\
    CW. rotation  (large-scale)  &  $\theta=-16$  \\
    \bottomrule
    \label{tab:rotation_paras}
    \end{tabularx}
    }
\end{table}


\section{Related Proofs}
\label{sec:proofs}
\textbf{Theorem 1.} 
With a fixed camera center, the result of point cloud rotation is independent of depth information.

\textit{Proof.}
Let $(u, v, 1)^{T}$ be the pixel coordinates in the original latent space, $K$ the camera intrinsic matrix, and $(X_c, Y_c, d(u,v,1))^{T}$ the spatial coordinates after point cloud projection, with $d(u,v,1)$ representing the depth.
Through the pin-hole camera model, we have:
\begin{equation}
    d(u,v,1) \left(\begin{array}{l}
    u \\
    v \\
    1
    \end{array}\right)=K \cdot\left(\begin{array}{l}
    X_{c} \\
    Y_{c} \\
    d(u,v,1)
    \end{array}\right)=\left(\begin{array}{ccc}
    f_{x} & 0 & c_{x} \\
    0 & f_{y} & c_{y} \\
    0 & 0 & 1
    \end{array}\right) \cdot\left(\begin{array}{l}
    X_{c} \\
    Y_{c} \\
    d(u,v,1)
    \end{array}\right),
\label{eq:11}
\end{equation}
where $f_x$and $f_y$ are the focal lengths, and $c_x$ and $c_y$ are the coordinates of the camera's optical center.

Rearranging the above equations, we obtain:
\begin{equation}
    \left\{\begin{array} { l } 
{ u = f _ { x } \cdot \frac { X _ { c } } { d(u,v,1) } + c _ { x } } \\
{ v = f _ { y } \cdot \frac { Y _ { c } } { d(u,v,1) } + c _ { y } }
\end{array} \Rightarrow \left\{\begin{array}{l}
X_{c}=\frac{\left(u-c_{x}\right)}{f_{x}} \cdot d(u,v,1) \\
Y_{c}=\frac{\left(v-c_{y}\right.}{f_{y}} \cdot d(u,v,1)
\end{array}\right.\right. .
\label{eq:15}
\end{equation}

According to the point cloud projection theory, by rotating the point cloud to another perspective using the rotation matrix $R_y$ (taking the rotation around the $Y$-axis as an example, with the same principle applying to rotations around other axes), we obtain the following equation:
\begin{equation}
    \left(\begin{array}{l}
    X^{\prime} \\
    Y^{\prime} \\
    Z^{\prime}
    \end{array}\right)=
    R_y \cdot\left(\begin{array}{l}
    X_{c} \\
    Y_{c} \\
    d(u,v,1)
    \end{array}\right)
    =\left(\begin{array}{ccc}
    \cos \theta & 0 & \sin \theta \\
    0 & 1 & 0 \\
    -\sin \theta & 0 & \cos \theta
    \end{array}\right) \cdot\left(\begin{array}{l}
    X_{c} \\
    Y_{c} \\
    d(u,v,1)
    \end{array}\right) .
\end{equation}

\begin{figure*}[!t]
\begin{center}
   \includegraphics[width=\textwidth]{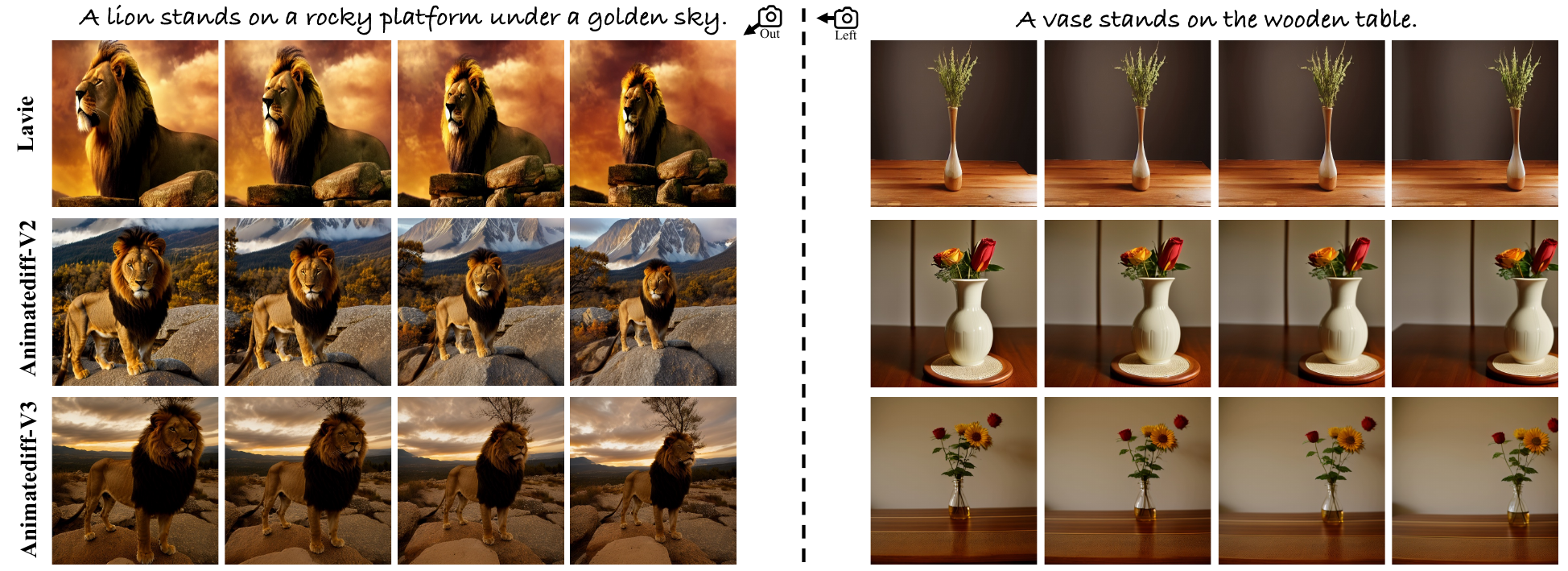}
   \caption{
   Our \method can be integrated into most existing frameworks, further demonstrating its scalability.
   }
   \label{fig:compatibility}
\end{center}
\end{figure*}

\begin{figure*}[!t]
\begin{center}
   \includegraphics[width=\textwidth]{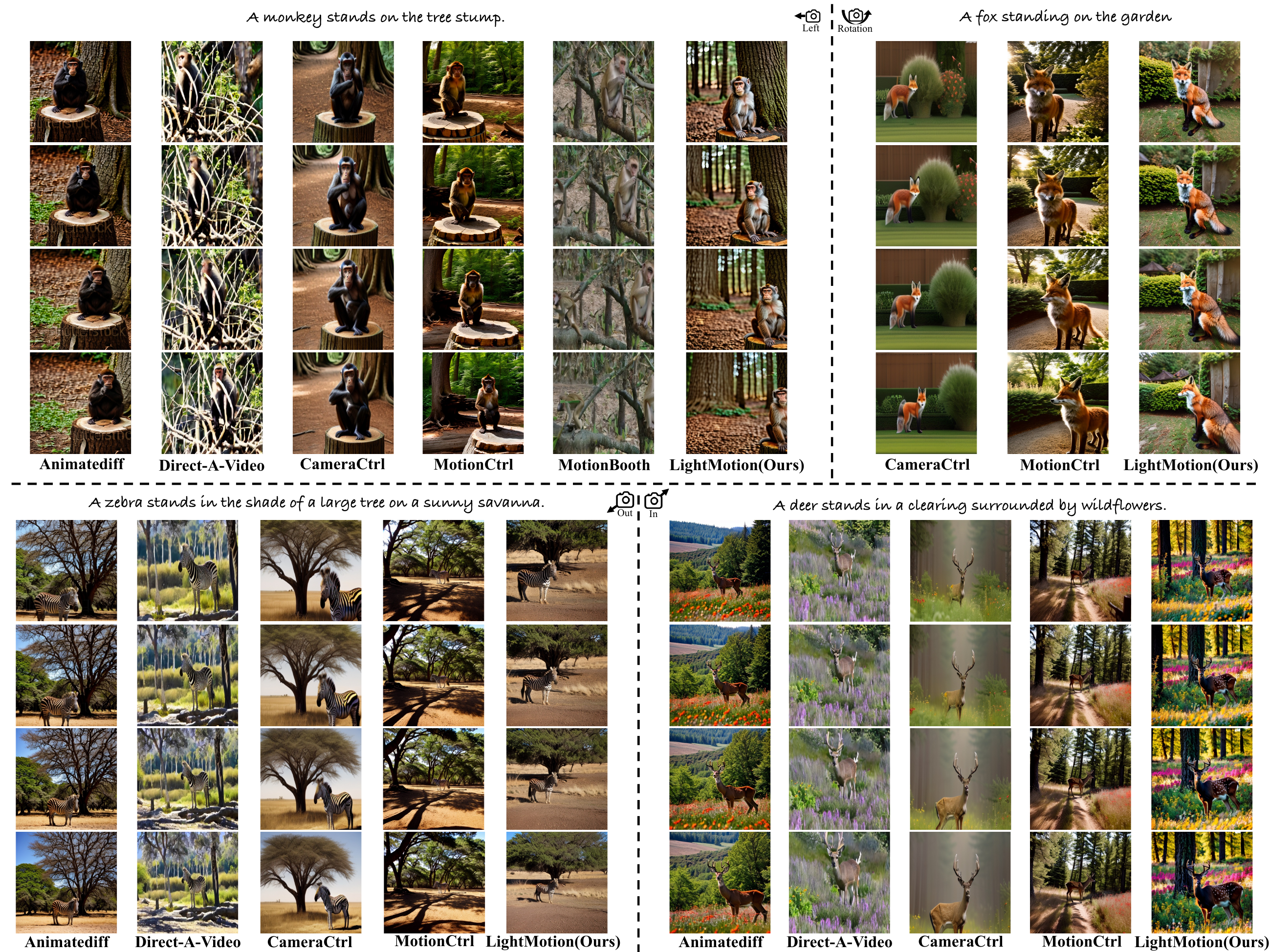}
   \caption{More qualitative comparisons with existing methods. For each camera motion, we compare only those methods that support it.
   }
   \label{fig:addition1}
\end{center}
\end{figure*}

\begin{figure*}[!t]
\begin{center}
   \includegraphics[width=\textwidth]{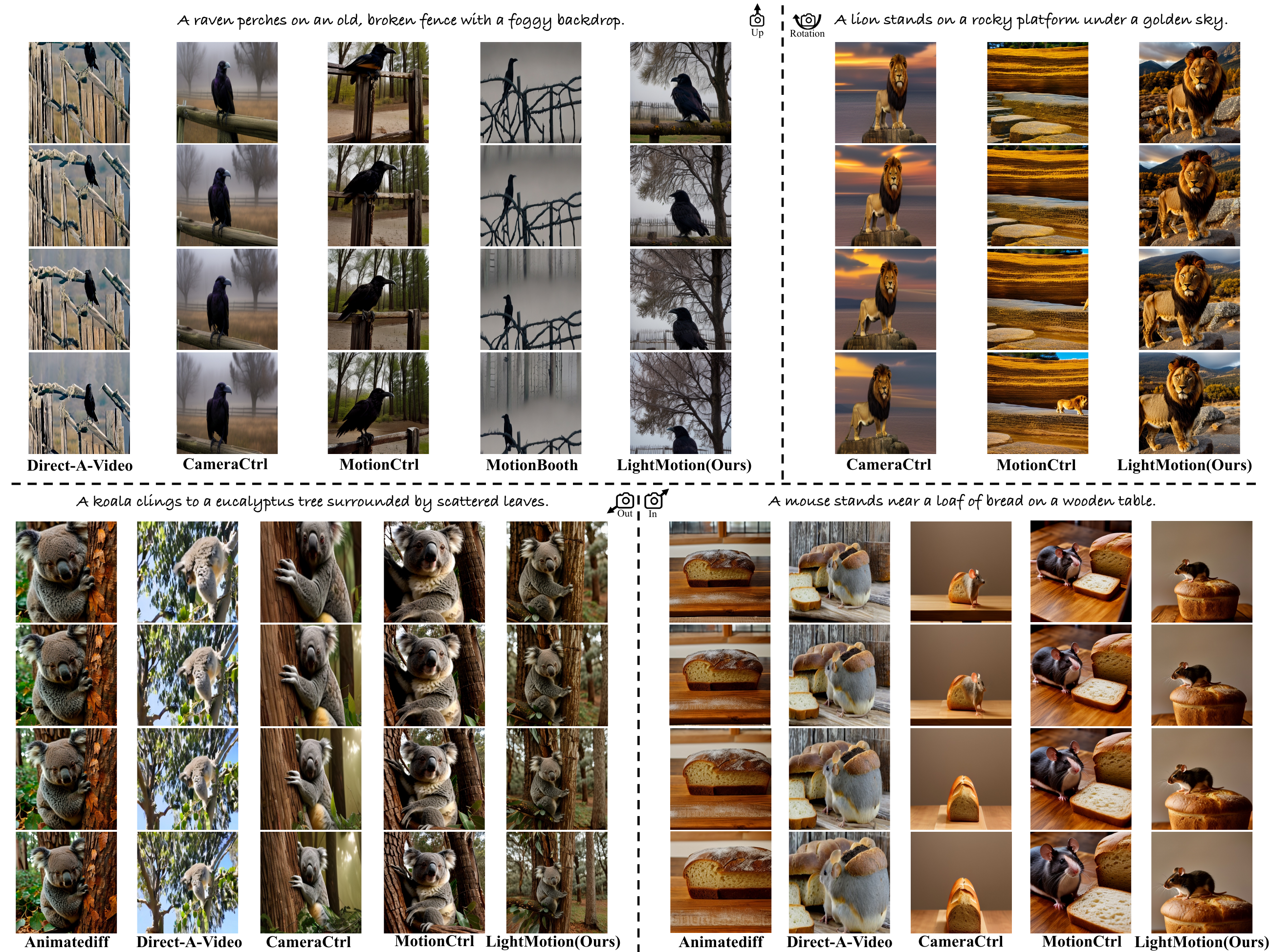}
   \caption{More qualitative comparisons with existing methods. For each camera motion, we compare only those methods that support it.
   %
   }
   \label{fig:addition2}
\end{center}
\end{figure*}

Substituting Eq.\,(\ref{eq:15}) and simplifying, we have:
\begin{equation}
    \left(\begin{array}{c}
    X^{\prime} \\
    Y^{\prime} \\
    Z^{\prime}
    \end{array}\right)=\left(\begin{array}{c}
    \cos \theta \cdot X_{c}+\sin \theta \cdot d(u,v,1) \\
    Y_{c} \\
    -\sin \theta \cdot X_{c}+\cos \theta \cdot d(u,v,1)
    \end{array}\right)=\left(\begin{array}{c}
    \frac{\cos \theta \cdot d(u,v,1) \cdot\left(u-c_{x}\right)}{f_{x}}+\sin \theta \cdot d(u,v,1) \\
    \frac{\left(v-c_{y}\right)}{f_{y}} \cdot d(u,v,1) \\
    \frac{-\sin \theta \cdot d(u,v,1) \cdot(v-(y)}{f_{y}}+\cos \theta \cdot d(u,v,1)
    \end{array}\right) .
\end{equation}

Then, we can derive the following ratio relationship:
\begin{equation}
   \begin{aligned}
   \left\{
   \begin{array}{ll}
\frac{X^{\prime}}{Z^{\prime}} &= 
\frac{\cos \theta \cdot f_{y} \cdot \left (u-c_{x}\right) + \sin \theta \cdot f_{x} \cdot f_{y}}
{-\sin \theta \cdot f_{x} \cdot \left(v-c_{y}\right) + \cos \theta \cdot f_{x} \cdot f_{y}} \\
\frac{Y^{\prime}}{Z^{\prime}} &= 
\frac{\left(v-c_{y}\right)}
{-\sin \theta \cdot \left(v-c_{y}\right) + \cos \theta \cdot f_{y}}
   \end{array} .
   \right.
   \end{aligned}
\label{eq:18}
\end{equation}

On the other hand, the point cloud in the new perspective can be mapped to the new pixel coordinates $(u',v',1)$ as in Eq.\,(\ref{eq:11}), satisfying the following relationship:
\begin{equation}
    Z^{\prime}\left(\begin{array}{l}
u^{\prime} \\
v^{\prime} \\
1
\end{array}\right)=K \cdot\left(\begin{array}{l}
X^{\prime} \\
Y^{\prime} \\
Z^{\prime}
\end{array}\right)=\left(\begin{array}{ccc}
f_{x} & 0 & C_{x} \\
0 & f_{y} & C_{y} \\
0 & 0 & 1
\end{array}\right) \cdot\left(\begin{array}{l}
X^{\prime} \\
Y^{\prime} \\
Z^{\prime}
\end{array}\right) .
\end{equation}

Substituting Eq.\,(\ref{eq:18}) and simplifying, we obtain:
\begin{equation}
    \left\{\begin{array}{l}
u^{\prime}=f_{x} \cdot \frac{X^{\prime}}{Z^{\prime}} + c_{x}=\frac{\cos \theta \cdot f_{y}\left(u-c_{x}\right)+\sin \theta f_{x} \cdot f_{y}}{-\sin \theta \cdot \left(v-c_{y}\right)+\cos \theta \cdot f_{y}}+c_{x} \\
v^{\prime}=f_{y} \cdot \frac{Y^{\prime}}{Z^{\prime}}+c_{y}=\frac{f_{y} \cdot\left(v-c_{y}\right)}{-\sin \theta\left(v-c_{y}\right)+\cos \theta \cdot f_{y}} + c_{y}
\end{array} . \right .
\end{equation}

The results show that projected pixel coordinates are independent of depth information $d(u,v,1)$. 
\textit{Proof End.}

\section{Compatibility with Different Frameworks}
\label{sec:compatibility}
Besides Animatediff-V2~\cite{AnimateDiff}, our method can also be seamlessly integrated into other video generation model frameworks like Animatediff-V3 and Lavie~\cite{Lavie}, with the additional qualitative results shown in Figure\,\ref{fig:compatibility}.

\section{Additional Qualitative Comparison}
\label{sec:additional_qualitative}
To highlight the superiority of \method, we provide additional qualitative comparisons in Figures\,\ref{fig:addition1} and \ref{fig:addition2}, showing its performance across various camera motions.

\section{Various Camera Combinations}
\label{sec:combinations}
\begin{figure*}[!t]
\begin{center}
   \includegraphics[width=\textwidth]{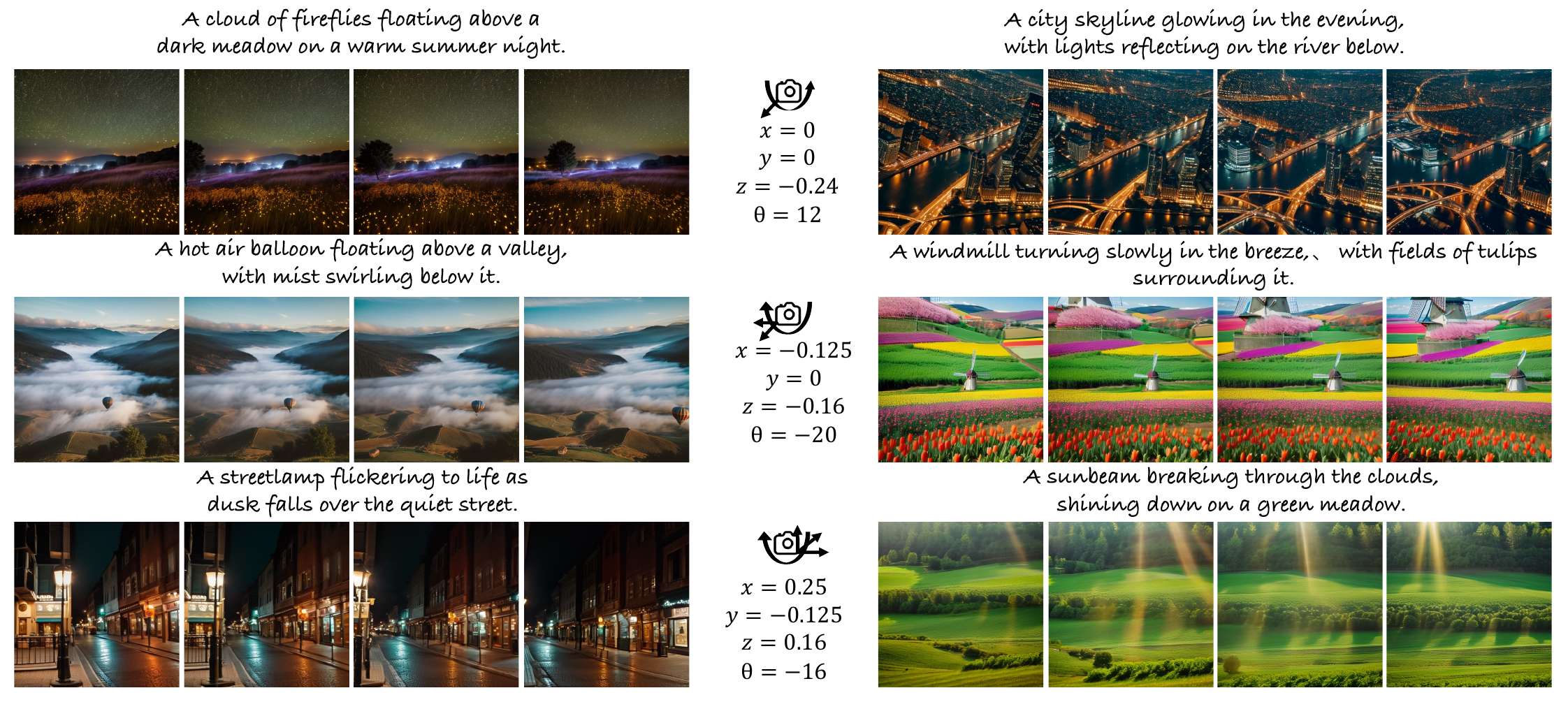}
   \caption{
   Additional video generation result with camera motion through user-defined parameter combinations.
   }
   \label{fig:combination1}
\end{center}
\end{figure*}

\begin{figure*}[!t]
\begin{center}
   \includegraphics[width=\textwidth]{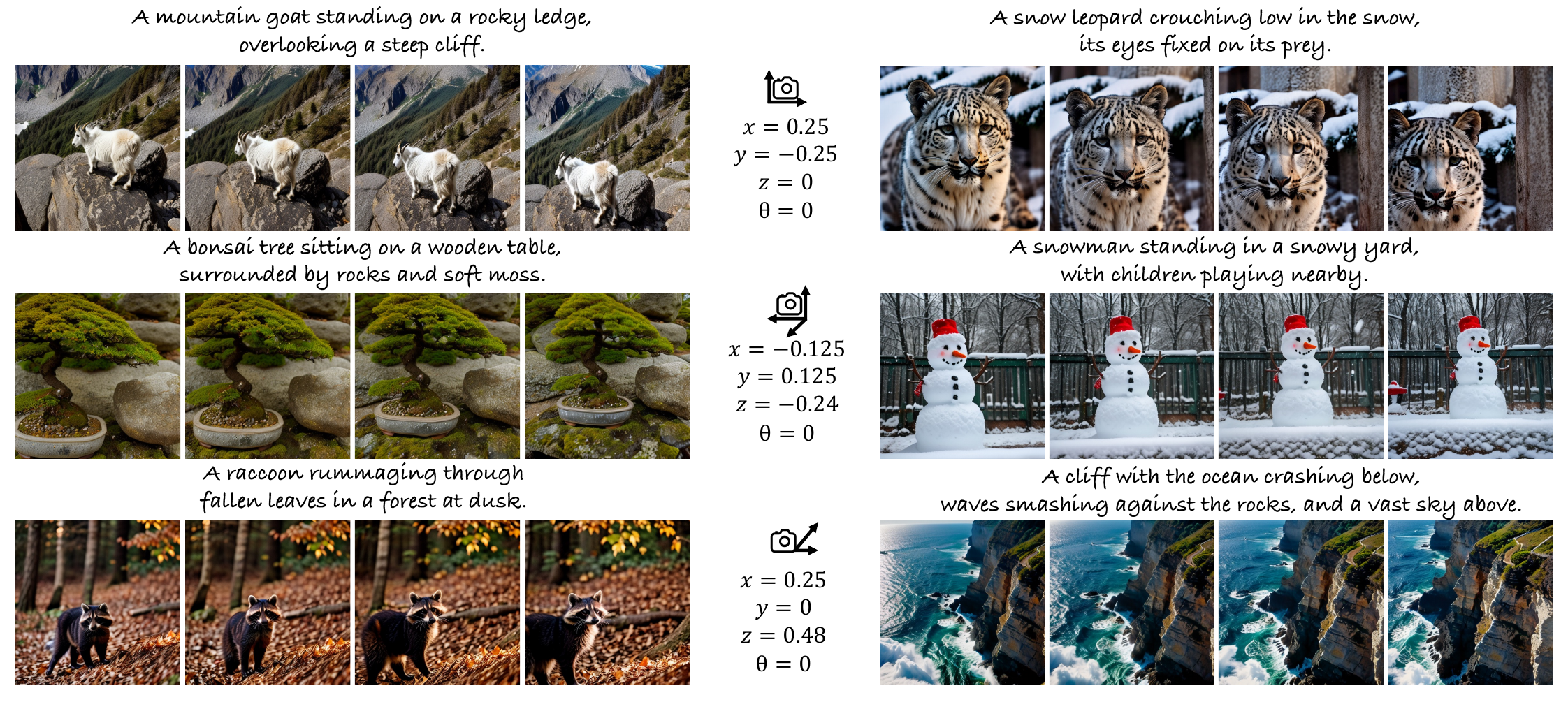}
   \caption{
   Additional video generation result with camera motion through user-defined parameter combinations.
   }
   \label{fig:combination2}
\end{center}
\end{figure*}

Furthermore, our \method supports a wide variety of camera combinations, with additional visual results provided in Figures\,\ref{fig:combination1} and \ref{fig:combination2}, demonstrating its versatility.

\section{User Study}
\label{sec:user_study}
We further assess the preferences of users for various methods through a user study.
Specifically, we design a questionnaire that includes 10 sets of videos generated by different methods. 
These sets include two groups featuring camera panning, four groups focusing on camera zooming, and four groups highlighting camera rotation.
Additionally, each generated video is accompanied by a relevant text description and the corresponding camera motion.
In each set, the results from all methods are transformed into .gif files and presented on same page.
We establish clear evaluation criteria for users, who score the videos on a scale of up to 100 in each set based on the following two aspects:
(i) \textit{Generation Quality}: This criterion evaluates the similarity between the generated video and its text description, as well as the aesthetic quality.
(ii) \textit{Camera Controllability}: This criterion assesses the alignment between the camera movements in the generated video and the specified camera motions.
To ensure fairness, the names of all methods will be concealed, and the order of the generated results in each set will be randomized.
Finally, 100 valid questionnaires were included in the analysis to evaluate user preferences.

\section{GPT-4o Evaluation}
\label{sec:gpt4o}
Additionally, the generated video samples for the user study will also be re-evaluated by GPT-4o.
Similarly, we establish clear evaluation criteria for GPT-4o, which scores the videos on a scale of up to 100 in each set based on the following three aspects:
(i) \textit{Generation Quality}: This criterion evaluates the similarity between the generated video and its text description, as well as the aesthetic quality.
(ii) \textit{Coherence}: This criterion evaluates the semantic coherence between frames in the generated video.
(iii) \textit{Camera Controllability}: This criterion assesses the alignment between the camera movements in the generated video and the specified camera motions.
To ensure robustness, we perform five repetitions and average the scores for each method.

\end{document}